\documentclass[letterpaper]{article} 
\usepackage{aaai24}  
\usepackage{times}  
\usepackage{helvet}  
\usepackage{courier}  
\usepackage[hyphens]{url}  
\usepackage{graphicx} 
\usepackage{natbib}  
\usepackage{caption} 
\usepackage[ruled,vlined,linesnumbered]{algorithm2e}

\usepackage{booktabs}
\usepackage{multirow}
\usepackage{amsmath}
\usepackage{amssymb}
\usepackage{mathtools}
\usepackage[capitalize]{cleveref}
\crefname{section}{Sec.}{Secs.}
\Crefname{section}{Section}{Sections}
\Crefname{table}{Table}{Tables}
\crefname{table}{Tab.}{Tabs.}

\title{Language-Guided Transformer for Federated Multi-Label Classification}
\author{
    I-Jieh Liu\textsuperscript{\rm 1},
    Ci-Siang Lin\textsuperscript{\rm 1,2},
    Fu-En Yang\textsuperscript{\rm 1,2},
    Yu-Chiang Frank Wang\textsuperscript{\rm 1,2}
}
\affiliations{
    \textsuperscript{\rm 1}Graduate Institute of Communication Engineering, National Taiwan University \ 
    \textsuperscript{\rm 2}NVIDIA\\
    \texttt{\small \{r11942087, d08942011, f07942077\}@ntu.edu.tw} \ \texttt{\small frankwang@nvidia.com}\\
    
}

\begin{document}

\maketitle
\begin{abstract}
Federated Learning (FL) is an emerging paradigm that enables multiple users to collaboratively train a robust model in a privacy-preserving manner without sharing their private data. Most existing approaches of FL only consider traditional single-label image classification, ignoring the impact when transferring the task to multi-label image classification. Nevertheless, it is still challenging for FL to deal with user heterogeneity in their local data distribution in the real-world FL scenario, and this issue becomes even more severe in multi-label image classification. Inspired by the recent success of Transformers in centralized settings, we propose a novel FL framework for multi-label classification. Since partial label correlation may be observed by local clients during training, direct aggregation of locally updated models would not produce satisfactory performances. Thus, we propose a novel FL framework of \textbf{L}anguage-\textbf{G}uided \textbf{T}ransformer (\textbf{FedLGT}) to tackle this challenging task, which aims to exploit and transfer knowledge across different clients for learning a robust global model. Through extensive experiments on various multi-label datasets (e.g., FLAIR, MS-COCO, etc.), we show that our FedLGT is able to achieve satisfactory performance and outperforms standard FL techniques under multi-label FL scenarios. Code is available at \url{https://github.com/Jack24658735/FedLGT}.
\end{abstract}

\section{Introduction}\label{sec:intro}

\begin{figure}[t]
	\centering
	\includegraphics[width=0.93\columnwidth]{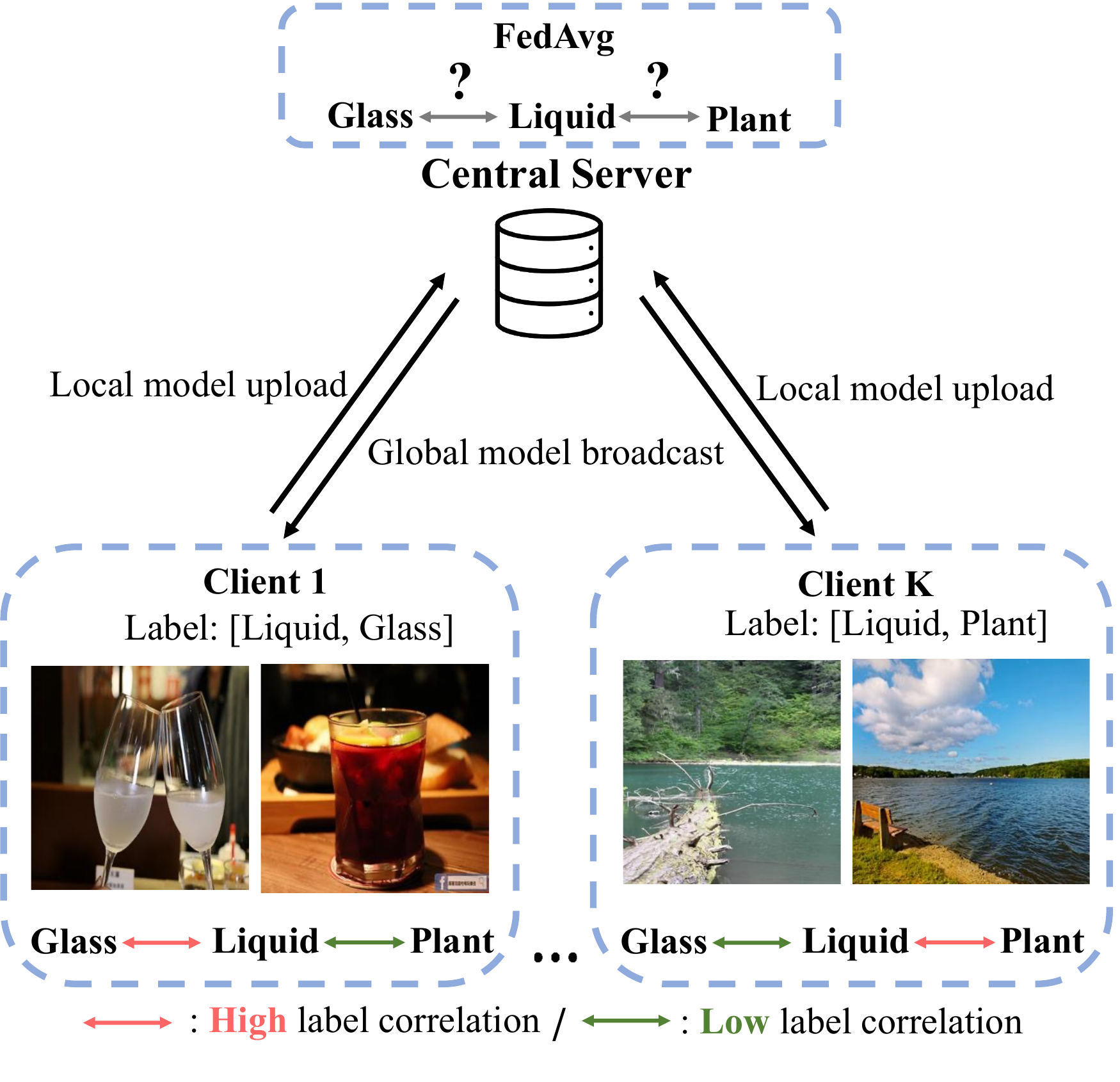}
    \caption{\textbf{Challenges in multi-label federated learning.} Since diverse label correlations can be observed across clients, aggregating local models might not be sufficiently generalizable.}
	\label{fig:intro}
\end{figure}

Federated Learning (FL) is a machine learning paradigm that enables multiple clients to collaboratively perform model training by leveraging their decentralized data. This approach offers significant advantages over traditional centralized approaches, as it mitigates the risks of privacy disclosure and is efficient for large-scale machine learning tasks without collecting large amounts of data on the server. FL has been successfully applied in real-world studies including health care, finance, and recommendation systems. As a pioneering work, the vanilla FL algorithm FedAvg~\cite{mcmahan2017communication} learns from multiple local clients in a privacy-preserving manner, and the trained models or gradients are sent to the central server. Then, the server performed an aggregation step to obtain a global model. However, user data distribution often has non-IID (non-independent and identical distribution) characteristics in the real world, which is referred to as \textit{data heterogeneity}. Data heterogeneity across different clients can be caused by various factors, such as demographic diversity, data collection methods, and sensor differences. In the presence of data heterogeneity, FedAvg may struggle to effectively leverage the data diversity across clients, resulting in degraded performance of the aggregated global model. 

Typically, data heterogeneity contains the aspects such as label distribution skew or domain shift, presenting substantial obstacles in the progress of FL development.
For instance, consider a scene understanding task in FL. In this scenario, certain clients possess a larger volume of indoor scene data, while others may have more outdoor scene data. As a result, their local models tend to be biased towards their respective scenes due to environmental variations in lighting, texture, etc., posing challenges in achieving a consensus during the model aggregation process (i.e., hard to obtain a robust global model). In addition, some clients may have more data for specific classes, resulting in a class imbalanced issue that can also cause the performance degradation of the aggregated global model. Prior works~\cite{li2020federated,karimireddy2020scaffold,gao2022feddc,li2021model,li2021fedbn} tackle data heterogeneity mainly on standard single-label image classification benchmarking datasets, e.g., EMNIST~\cite{cohen2017emnist}, CIFAR-10~\cite{krizhevsky2009learning}, etc. However, multi-label image classification is a more practical and challenging setting. For example, single-label image classification usually needs the model to recognize only one object or one main concept in an input image, whereas multi-label image classification aims to recognize all the object categories or concepts, which is more challenging due to the understanding of the inter-class relationships. 

To mitigate performance degradation induced by data heterogeneity (e.g., label distribution skew) on FL, various works~\cite{li2020federated,karimireddy2020scaffold,gao2022feddc,li2021model,li2021fedbn,huang2022learn,su2022cross,mendieta2022local,wang2020tackling, tan2021fedproto,zhuang2021collaborative,luo2021no} have been proposed to address this impact. For example, FedProx~\cite{li2020federated} introduces a regularization term to mitigate the label distribution skew issues from the local learning step of FL. On the other hand, a previous FL method for addressing domain shift called FedBN~\cite{li2021fedbn} proposes utilizing the batch normalization layer~\cite{ioffe2015batch} in local clients to capture the domain-specific information. However, these FL previous works neglect the emerging form of data heterogeneity such as multi-label issues but only focus on traditional single-label non-IID issues, leaving the \textit{``multi-label FL"} task to remain unresolved and challenging.

Intuitively, it is achievable to directly apply the current multi-label classification frameworks from centralized learning~\cite{chen2019multi,chen2019learning,chen2020knowledge,lanchantin2020general} to FL. However, when performing multi-label classification under FL, the objects or concepts present in an image may exhibit significant variations across clients, potentially causing performance deterioration. For example, as~\Cref{fig:intro} depicted, the positive label set of each client is inconsistent or even disjoint, learning of the label correlation may have unfavorable impacts and degrade the global model performance after the aggregation step. Thus, one of the main challenges for multi-label image classification in FL is how to effectively learn label relationships, as well as precisely capture the complex connections between visual features and associated labels in a \textit{privacy-preserving} manner.

In this paper, we propose a framework called \textit{FedLGT} that aims to tackle the multi-label FL issues and verify our effectiveness on a challenging multi-label FL dataset~\cite{song2022flair} called FLAIR. Our FedLGT utilizes a pre-trained off-the-shelf text encoder by CLIP~\cite{radford2021learning} to construct \textit{universal label embedding}. The label embedding contains rich and distinct relationships among the labels. Utilizing this technique enables the clients to perform local learning with more discriminative label relationships. Besides, with the goal of training a robust global model from decentralized data, we also design a knowledge-transfer approach called \textit{Client-Aware Masked Label Embedding} inspired by~\cite{lanchantin2020general} to assist local learning via knowledge from the global model. 
Specifically, the approach aims to transfer and distill the knowledge from the global model by encouraging the local models to learn more for the classes that the global model still predicts with relatively low confidence. Through conducting extensive experiments on FLAIR, our proposed framework is able to derive a global model with better generalization performance without any communication overheads.

Our contributions to this work are highlighted as follows:
\begin{itemize}
    \item To the best of our knowledge, we are the first to tackle the problem of label discrepancy across different clients for multi-label FL.
    \item We propose \textit{Client-Aware Masked Label Embedding} when training FedLGT. It is served as a customized model update technique while exploiting the label correlation at each client.
    \item We utilize \textit{Universal Label Embedding} in FedLGT, which advances pre-trained label embedding derived from large-scale vision and language models (e.g., CLIP) for aligning local models for multi-label FL.
\end{itemize}

\section{Related works}\label{sec:related}

\subsection{Federated Learning}
In general, the common vanilla FL algorithm (i.e., FedAvg~\cite{mcmahan2017communication}) consists of several steps including local training, uploading client models, performing model aggregation, and broadcasting back to the clients on the server side. Under such FedAvg training procedures, the inconsistency between local objectives and the global objective would be more serious under data heterogeneity (e.g., domain shift). 
The main direction of existing previous methods could be broadly divided into \textit{label distribution skew} or \textit{domain shift}. Label distribution skew can arise due to disparate local data in different clients, leading to bias local models towards majority classes. On the other hand, domain shift does not specifically highlight the differences in label distributions across clients. Instead, the image features may exhibit significant variations, such as cartoons, sketches, etc., even under the same label annotation. In the subsequent sections, we mainly introduce the works that are popular or more relevant to our paper.

\paragraph{Label Distribution Skew}
Label distribution skew is a kind of common challenge in FL. For example, assume that some hospitals want to train a model to classify medical images. However, one client may have some rare diseases, while other clients have other common diseases. Thus, the label distribution will differ significantly or even disjoint (i.e., label spaces are not overlapped), leading to degradation of model performance. In terms of label distribution skew, existing FL methods focus on handling client local bias compared to the global model. For instance, FedProx~\cite{li2020federated} restricts the gradient updates by introducing a proximal term to improve the convergence speed. SCAFFOLD~\cite{karimireddy2020scaffold} designs a new control variate for each client by measuring the gradient dissimilarity between local clients and the global model, utilizing this mechanism to refine the local clients' drift. More recently, FedDC~\cite{gao2022feddc} learns an auxiliary local drift variable by Expectation-Maximum (EM) algorithm to track the inconsistency of local-global models. 

\paragraph{Domain Shift}
Domain shift (i.e., distribution shift) is another challenge among many real-world FL applications. For instance, a healthcare organization wants to perform training on patient data of certain diseases (i.e., the label distributions are the same) from different hospitals. Each of them maintains its own private dataset of medical images which may have different characteristics due to variations in the data acquisition process, patient populations, etc., resulting in potential domain shifts. Several previous works have been proposed to mitigate the domain shift of FL~\cite{li2021fedbn,huang2022learn,su2022cross}. FedBN~\cite{li2021fedbn} does not aggregate the parameters of batch normalization (BN) layers, so the domain-specific information can be preserved in local clients. FCCL~\cite{huang2022learn} builds a cross-correlation matrix on the server with the help of unlabeled public data. Besides, it balances the inter-domain and intra-domain information in the local training stage to tackle the domain shift. Recent FL works~\cite{su2022cross, guo2023promptfl, chen2022fedtune, sun2022exploring} utilize vision-language pre-trained models, which perform prompt tuning during training rounds and thus reduce the number of learnable parameters. FedAPT~\cite{su2022cross} learns a global adaptive network with the global prompt under the FL setting, then the framework generates domain-specific prompts for CLIP to handle the domain shift issue under FL. Inspired by this trend, our approach leverages vision-language pre-trained models (e.g., CLIP) to circumvent the challenges of training components that could deteriorate the model performance under the constraints of FL.

\subsection{Multi-label Image Classification}
\subsubsection{Centralized Learning}\label{ssec:cent_ml}
Multi-label image classification has become an emerging research field since several realistic applications may view this task as a foundation, such as weakly supervised segmentation, scene understanding, etc. Since multi-label image classification requires recognition of all objects or concepts present in an image, it is critical yet challenging to understand the inherent relationships between different classes. There are many directions for multi-label image classification, such as improving loss functions~\cite{ridnik2021asymmetric}, modeling label relationships~\cite{chen2019multi,lanchantin2020general}, etc. The most related to our work is how to model the label relationships so that co-occurrence dependencies between different classes would be considered appropriately. Specifically, recent methods~\cite{chen2019multi,chen2019learning,chen2020knowledge} utilize graph formulation to represent label relationships. For example, ML-GCN~\cite{chen2019multi} build \textit{Graph
Convolutional Network (GCN)} to represent the correlation between objects or concepts, and such a method usually builds label correlation graphs based on extra knowledge from label co-occurrence statistics.

Based on Transformer, C-Tran~\cite{lanchantin2020general} fuses image features and the associated labels by \textit{label mask training}. Similar to BERT~\cite{devlin2018bert}, C-Tran aims to capture semantic information by predicting the unknown labels on the images with the help of known labels. However, in FL settings, label co-occurrence information might differ across distinct clients. As confirmed later by our experiments, applying standard FL techniques with C-Tran might not be preferable.

\subsubsection{Federated Learning}\label{ssec:fed_ml}
Intuitively, one can utilize centralized multi-label learning models in FL schemes. That is, one can view each local client as a multi-label classification task and perform aggregation on the server side. However, existing centralized approaches require interaction with training data to capture a global view of the label relationships, which is infeasible under FL scenarios. For example in Fig.1, one client primarily captures images of drinks and glass, while the other client mainly takes pictures of landscapes and plants. Thus, the former client's model learns a high correlation between ``liquid" and ``glass", whereas the latter client's model learns a high correlation between ``liquid" and ``plant". However, aggregating these models on the server side could lead to confusion between these categories (i.e., ``liquid", ``glass", and ``plant") for the global model, resulting in 
a degradation of performance. 

To the best of our knowledge, recent FL works did not address multi-label learning tasks, and thus the above issue would limit the model performances. As discussed in the following section, we propose a \textit{Transformer-based} model while tackling the inherent label distribution differences between clients. Compared to popular FL techniques like FedAvg, our method achieves impressive results on a large-scale dataset for multi-label FL (i.e., FLAIR~\cite{song2022flair}).

\begin{figure*}[t]
	\centering
	\includegraphics[width=1.95\columnwidth]{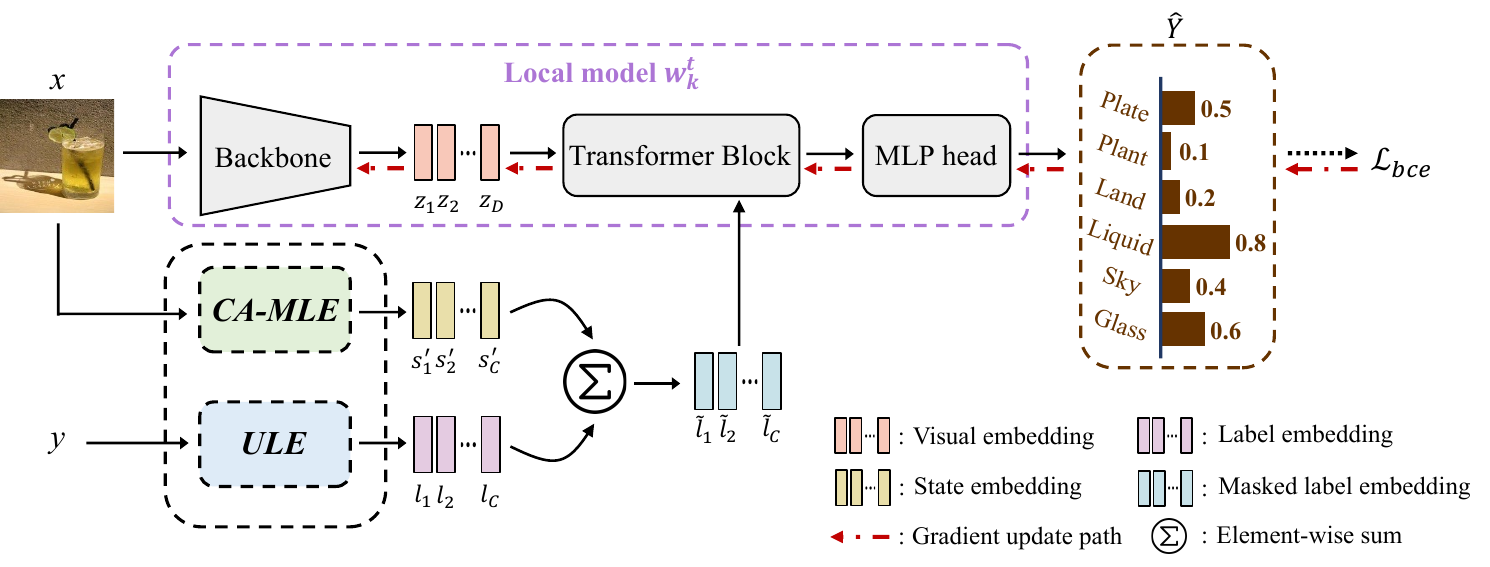}
    \caption{\textbf{Overview of FedLGT.} Given an image with multi-labels to predict, the global model from each communication round updates the local model with Client-Aware Masked Label Embedding (CA-MLE), which exploits partial label correlation observed at each client. In order to properly align local models for multi-label FL, universal label embeddings (ULE) are utilized in FedLGT. (Best viewed in color.)
    }
	\label{fig:archi}
\end{figure*}

\section{Proposed Method} \label{sec:method}

\subsection{Problem Formulation and Setup}
In this work, we assume that $K$ clients would be involved in the federated learning process in each communication round, and local private datasets are available for clients $D = \{D_1, D_2, ..., D_K\}$ during training. For each client, it can be viewed as solving a standard multi-label image classification task. To be more specific, given an image $x$ with the corresponding label $y=[y_1, ...,y_C]$, where $C$ is the number of classes, $y_i=1$ indicates $i$-th class is present in the image, and $y_i=0$ is not present, then the local clients attempt to predict the existence of each category of the image.
The goal is to obtain a global aggregated model handling multi-label classification on FL that solves the objective
\begin{equation}\label{eq:obj_func}
    w^* = \mathop{\arg\min}_w \Sigma_{k=1}^{K}\frac{|D_k|}{|D|}\mathcal{L}_k(w),
\end{equation}

In multi-label FL, each client and server share the same label space with a total of $C$ categories. However, label distributions might differ across different clients. Thus, based on a recent SOTA centralized multi-label learning model of C-Tran~\cite{lanchantin2020general}, our goal is to tackle the multi-label FL task.

\subsection{A Brief Review for C-Tran}\label{ssec:c-tran intro}
C-Tran~\cite{lanchantin2020general} is a centralized \textit{Transformer-based}~\cite{vaswani2017attention} multi-label image classification framework, which is designed to observe image features and label correlations simultaneously. In C-Tran, the image features are extracted by ResNet~\cite{he2016deep}, while the labels are described by both label $L$ and state $S$ embeddings. The label embeddings are defined as $L=\{l_1, l_2, ..., l_C\}$, where each $l_c \in \mathbb{R}^d$ represents $c$-th class labels ($d$ denotes the embedding dimension). On the other hand, a set of state embeddings $S=\{s_1, s_2, ..., s_C\}$ (with each $s_c \in \mathbb{R}^d$) are viewed as tokens, indicating the presence of the corresponding labels of \textit{unknown}, \textit{positive}, and \textit{negative}. Note the encoded token value for the associated state embeddings is $-1$, $1$, and $0$, respectively. Also, only the \textit{unknown} state would contribute the loss to the model during training. With label and state embeddings, C-Tran proposes a training pipeline of \textit{Label Mask Training (LMT)} that randomly masks partial amounts of labels and has the model perform prediction, implicitly exploiting label correlations. Specifically, the state embeddings would be added to the aforementioned label embeddings to form the masked label embeddings $\tilde{l_c}$ for LMT: 
\begin{equation}\label{eq:label_emb}
    \tilde{l_c} = l_c + s_c,
\end{equation}
where $s_c$ is one of the states among \textit{unknown}, \textit{positive}, \textit{negative}. Thus, the masked label embeddings can be formulated as $\tilde{L}=\{\tilde{l}_1, \tilde{l}_2, ..., \tilde{l}_C\}$ in the LMT process.

As shown in Fig.~\ref{fig:archi}, the image features $Z$ extracted by the vision backbone (e.g., ResNet~\cite{he2016deep}) would be concatenated with the masked label embeddings $\tilde{L}$, and sent into the transformer model. Thus, one has
\begin{equation}\label{eq:ctran}
    \hat{Y} = w(x, \tilde{L}),
\end{equation}
where $x$ denotes the input image, $\tilde{L}$ represents the masked label embeddings, and $\hat{Y}$ is the predicted logit ($w$ denotes the network model). However, as noted in~\Cref{sec:related}, C-Tran cannot easily preserve label co-occurrence across different clients in FL settings. Thus, how to extend C-Tran for FL multi-label classification remains a challenging task.

\subsection{Federated Language-Guided Transformer}\label{ssec:our work}
In multi-label FL, we aim to learn a global model that generalizes to different clients with potential label distribution skews and diverse label correlations. Instead of simply performing model aggregation (like FedAvg) with pure visual input which might lead to degraded multi-label classification performances, we extend C-Tran and propose a novel learning scheme of Federated Language-Guided Transformer (FedLGT). As depicted in Fig.~\ref{fig:archi}, we introduce model updating and feature embedding learning schemes for FedLGT, as presented below.

\subsubsection{Client-Aware Masked Label Embedding} \label{sssec:calmt}
To train global and client models in multi-label FL settings, how to tackle the domain shift or label skews across clients while jointly exploiting inherent label co-occurrence information is a challenging problem. At the $t$-th training round, each client applies the global model $w^t$ to output the prediction vector $P=\{p_1, p_2, ...,p_C\}$, where each of $p_c \in [0,1]$ represents a probability indicating the presence of the associated label. However, since the global model would not generalize to each client during training rounds, one would expect some class labels to be with lower confidence (i.e., $p_c$ is around $0.5$ instead of being close to $1$ or $0$). In such cases, a client-aware training strategy is necessary for updating the associate local model and calibrating the state embeddings $s_c$. 

\begin{figure}[t]
	\centering
	\includegraphics[width=0.95\columnwidth]{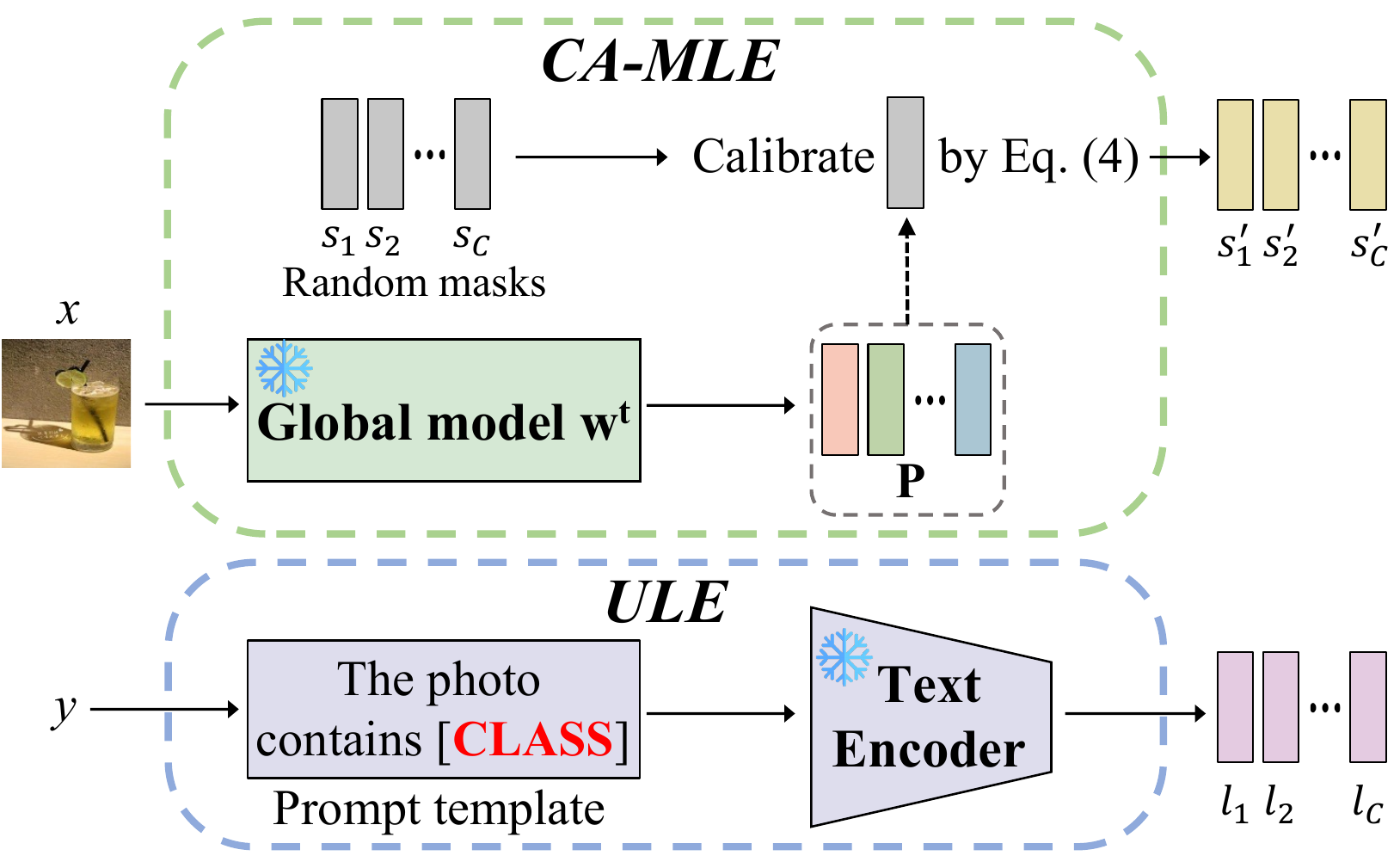}
    \caption{\textbf{Details of CA-MLE and ULE.} With the input image $x$ and label $y$, CA-MLE generates the prediction by global model $w^t$ and calibrates the state embeddings, while ULE advances the pre-trained label embeddings from CLIP for model aggregation purposes.
    }
    \label{fig:ule_calmt}
\end{figure}

To address this problem, we propose \textit{client-aware masked label embedding (CA-MLE)} as depicted in Fig.~\ref{fig:ule_calmt} for local model updating. In CA-MLE, we would modify the original state embeddings of $c$-th class to be \textit{unknown} if the $p_c$ is uncertain, otherwise, the state embeddings remain unchanged. More precisely, the calibrated state embeddings $S^{\prime}=\{s^{\prime}_1, s^{\prime}_2, ..., s^{\prime}_C\}$ are defined as 
\begin{equation}\label{eq:kd_2}
    s^{\prime}_c:
    \begin{cases}
     \text{\textit{unknown}},  & \tau - \varepsilon \leq p_c \leq \tau + \varepsilon  \\
     s_c, & \text{otherwise} \\
    \end{cases}
    ,
\end{equation}
where $\tau$ denotes the threshold (typically $0.5$ as set in most multi-label works~\cite{lanchantin2020general,liu2021query2label}) to determine the presence of the corresponding class, and $\varepsilon$ as the adaptable margin to describe the uncertainty. Note that only the unknown state would contribute losses to the training. Namely, if the probability lies in the interval, it indicates that the $c$-th class is uncertain for the global model $w^t$, and such a class is enforced to be further learned during that training round.

With calibrated state embeddings mentioned above, we view these embeddings as the masks to indicate which parts should be learned more during local training. We would not perform training by randomly generated masks as Eq.~\eqref{eq:label_emb} stated. Instead, we utilize these calibrated state embeddings $s^{\prime}_c$ to each of the label embeddings to form the masked label embeddings to enhance the local training.
Through the use of this calibration technique, we can harness the knowledge contained within the global model through state embeddings to guide the local training. Namely, it allows the local clients to focus more on those classes that are still challenging for the global model when local training, resulting in learning a better generalized and robust model after FL aggregation.

\subsubsection{Universal Label Embedding}
In federated multi-label classification, the aggregated global model would not properly capture label correlation across different clients, since it does not have access to data for each client. Recall that, for each client, the local models may only capture the partial view of label correlations. Since an aggregation step is commonly deployed in FL to update the global model, it would be desirable to design a proper local model alignment mechanism, so that the inherent label co-occurrence information observed at each client would be properly shared and aggregated at the server.

To accomplish this, as depicted in Fig.~\ref{fig:ule_calmt}, we propose to deploy \textit{universal label embedding (ULE)}  across clients in FedLGT. That is, we advance the large vision-language (VL) pre-trained model of CLIP~\cite{radford2021learning} and utilize its (fixed) text encoder for deriving pre-aligned and fixed label embeddings. To be more specific, we build a prompt \textit{``The photo contains [CLASS]"} as the input of the text encoder to generate the embedding for the associated $c$-th class label $l_c$. Such pre-trained/aligned label embeddings thus serve as the guidance during training of FedLGT, ensuring the updates of local models are aligned with pre-determined embedding space and thus favoring the subsequent model aggregation process.

\begin{algorithm}[t]
\caption{Training of FedLGT}\label{alg1}
\SetKwFunction{LocalUpdate}{\textbf{LocalUpdate}}
\SetKwFunction{ServerSide}{\textbf{ServerSide}}
\SetKwInOut{Parameter}{Parameter}

\KwIn{Communication rounds $T$, local epochs $E$, number of all clients $K$,  the fraction of active client in each round $\rho$, the initial model $w^0$, $k$-th client dataset $D_k$;}
\KwOut{The trained global model $w^t$}
\BlankLine
Build \textbf{universal label embedding} $L$ in Sec.~\ref{sssec:calmt}\\
\For {$t=1,2,...,T$} {
    Sample $R= \lceil \rho \cdot K \rceil$ clients to train \\ 
    \For {$k=1,2,...R$ \textbf{in parallel}}  {
        $w^t_k$ $\leftarrow$ \LocalUpdate{$w^t$, $L$} \\
        $w^{t+1} \leftarrow \Sigma_{k=1}^{K}\frac{|D_k|}{|D|}w^t_k$  \\
    }
}
return $w^t$

\BlankLine
\LocalUpdate{$w^t$, $L$} \\
\For {$e=1,2,...,E$} {
    $w^t_k$ $\leftarrow$ $w^t$ \\
    \For {($x_k, y_k$) $\in$ $D_k$} {
        $P$ $\leftarrow$ $w^t(x_k, y_k, L)$ \\
        Produce $\tilde{L}$ with $P$ by \textbf{CA-MLE} in Sec.~\ref{sssec:calmt} \\
        Update $w^t_k$ by $\mathcal{L}_{bce}$ \\
    }
}
return $w^t_k$

\end{algorithm}

\subsection{Training of FedLGT}
In FedLGT, we perform local model updates and global model aggregation in one communication round to achieve a federated learning process.
\paragraph{Local Model Update}
During the local model update, as depicted in ~\Cref{fig:archi}, each client $k$ would go through CA-MLE with ULE to produce masked label embeddings. 
For CA-MLE, the client would receive an aggregated global model from the previous round, and it would make a prediction on a given image to obtain the logits. With these logits from the global model, it can be served as guidance for the local model in the training process. It implies that the local model $w^t_k$ would identify the classes which are not properly described from the perspective of the global model. Thus, this allows the local model to focus more on those classes during training. As for ULE, the label correlations would be aligned according to fixed label embeddings across different clients. After passing these components, we follow Eq.~\eqref{eq:label_emb} to perform element-wise addition of these embeddings to form\textbf{ masked label embedding} to serve as the label inputs of the local model.

With masked label embedding, we leverage the self-attention mechanism in the transformer block to capture the relationships between the input image features and label correlations. To be more specific, we concatenate the masked label embeddings and image features extracted by the vision backbone (e.g., ResNet) as the input tokens of the transformer. Next, the output embeddings generated by the transformer are sent to the classifier  (i.e., MLP head) to produce the multi-label classification results (i.e., $\hat{Y}$ in ~\Cref{fig:archi}). 

As for the loss function to update the local model for client $k$ (i.e., $w^t_k$ shown in~\Cref{fig:archi}), we adopt binary cross-entropy loss with our adjustment for masks, and only the masked labels would be optimized during training, which is formulated as:
\begin{equation}\label{eq:bce_loss}
    \mathcal{L}_{bce}= -\sum_{\mathclap{\substack{s^{\prime}_c \text{is \textit{unknown}} \\
    \forall c\in [1, C]}}} y_c\log(p_c),
\end{equation}

\begin{table}[t]
  \centering
  \resizebox{1.0\columnwidth}{!}{
  \begin{tabular}{@{}lcccccccc@{}}
    \toprule
    Metrics & C-AP & C-P & C-R & C-F1 & O-AP & O-P & O-R & O-F1\\
    \midrule
    \multicolumn{8}{@{}l@{}}{\textit{\textbf{Centralized (upper bound)}}} \\
    ResNet & 67.71 & 75.71 & 55.42 & 64.00 & 90.40 & 84.09 & 78.96 & 81.44 \\
    C-Tran & 71.60 & 76.30 & 62.00 & 68.40 & 91.50 & 84.40 & 80.70 & 82.50 \\
    \midrule

    \multicolumn{8}{@{}l@{}}{\textit{\textbf{Federated}}} \\
    FedAvg& 40.63 & 42.74 & 26.53 & 32.74 & 77.39 & 78.35 & 56.31 & 65.53 \\
    FedC-Tran & 56.00  & 49.40  & 38.20  & 43.10  & 88.10  & 83.10  & 72.50  & 77.40  \\
    \midrule
    Ours & \textbf{60.90} & \textbf{67.80} & \textbf{46.50} & \textbf{55.10} & \textbf{88.70} & \textbf{84.00} & \textbf{75.90} & \textbf{79.70} \\
    \bottomrule
  \end{tabular}}
  \caption{Comparisons of \textbf{coarse-grained} multi-label classification task on FLAIR. \textbf{Bold} denotes the best result under the FL setting.}
  \label{tab:table_1}
\end{table}

\paragraph{Global Model Aggregation}
Once the local training is done, the clients would upload their local models $w^t_k$ to the server and perform the aggregation step in FL. Specifically, we utilize the vanilla FedAvg~\cite{mcmahan2017communication} aggregation process, and it can be derived as follows:
\begin{equation}\label{eq:agg}
    w^{t+1} = \Sigma_{k=1}^{K}\frac{|D_k|}{|D|}w^t_k\\,
\end{equation}
As soon as the training of our framework is converged, the trained aggregated global model is capable of performing multi-label image classification tasks by setting all the state embeddings to \textit{unknown} for prediction.

With the above learning process, we are able to transfer the knowledge of the global model to local clients, with inherent and partial label correlation properly shared and updated. The pseudo-code of our proposed framework is described in~\Cref{alg1}. Once the training of FedLGT is complete, we apply the learned model for testing.

\section{Experiments}\label{sec:exp}

\subsection{Experimental Setup}
\paragraph{Datasets}
To demonstrate the effectiveness of our proposed learning framework, we conduct extensive evaluations on various benchmark datasets, including FLAIR~\cite{song2022flair}, MS-COCO~\cite{lin2014microsoft}, and PASCAL VOC~\cite{everingham2015pascal}. Specifically, we primarily evaluate our method on the recently introduced multi-label FL dataset, FLAIR. FLAIR is a large-scale multi-label FL dataset, which contains a wide variety of photos collected from real users on Flickr. FLAIR provides real-user data partitions with each input image in $256 \times 256$ pixels. Thus, FLAIR naturally captures various non-IID characteristics, including quantity skew (i.e., users have different numbers of samples), label distribution skew, and domain shift, leading to a more challenging scenario for FL. FLAIR is defined in a two-level hierarchy, one task called \textbf{coarse-grained} with 17 categories, the other called \textbf{fine-grained} with 1,628 categories. Besides, we also validate the performance of our method on centralized datasets partitioned artificially for FL, including MS-COCO and PASCAL VOC. To the best of our knowledge, these datasets have not been explored before in the context of multi-label FL. MS-COCO contains about 122,218 images containing common objects, and the standard multi-label formulation covers 80 object class annotations for each image. Moreover, the images in PASCAL VOC include multiple labels, corresponding to 20 object categories, which is a relatively easier task than FLAIR and MS-COCO.

\begin{table}[t]
  \centering
  \resizebox{1.0\columnwidth}{!}{
  \begin{tabular}{@{}lrrrrrrrr@{}}
    \toprule
    Metrics & C-AP & C-P & C-R & C-F1 & O-AP & O-P & O-R & O-F1\\
    \midrule
    \multicolumn{8}{@{}l@{}}{\textit{\textbf{Centralized (upper bound)}}} \\
    ResNet & 20.26  & 32.97  & 10.92  & 16.40  & 47.95  & 68.73  & 30.04  & 41.81  \\
    C-Tran & 27.50  & 33.10  & 13.30  & 18.90  & 54.20  & 71.00  & 34.70  & 46.60  \\
    \midrule
    \multicolumn{8}{@{}l@{}}{\textit{\textbf{Federated}}} \\
    FedAvg & 2.03  & 1.99  & 0.40  & 0.66  & 27.31  & 65.47  & 10.50  & 18.10  \\
    FedC-Tran & 3.30  & 3.00  & 1.00  & 1.50  & 36.70  & 69.10  & 20.60  & 31.70  \\
    \midrule
    Ours & \textbf{10.60} & \textbf{6.50} & \textbf{1.40} & \textbf{2.30} & \textbf{42.20} & \textbf{69.80} & \textbf{21.90} & \textbf{33.40} \\
    \bottomrule
  \end{tabular}}
  \caption{Comparisons of \textbf{fine-grained} multi-label classification task on FLAIR. \textbf{Bold} denotes the best result under the FL setting. 
  }
  \label{tab:table_2}
\end{table}

\paragraph{Implementation Details}
We use ResNet-18~\cite{he2016deep} as our vision backbone across all datasets mentioned above. For the universal label embeddings, we
build a prompt \textit{``The photo contains [CLASS]”} as the input
of the text encoder for CLIP~\cite{radford2021learning}, whereas for the state embeddings, we build the prompt with one of the `positive" or `negative" for CLIP, and ``unknown" is fixed with all zeros. Note that for FLAIR coarse-grained level since the categories are abstract, we build the prompts in two ways via the fine-grained level information. One collects the fine-grained labels for each coarse-grained label to build the prompts (i.e., [CLASS] parts), and the other builds all embeddings of the fine-grained labels, and for each coarse-grained label, taking average on corresponding fine-grained to build the coarse level embeddings. As for CA-MLE, we set the threshold $\tau$ to $0.5$ and the uncertainty margin $\varepsilon$ is $0.02$. For each round of local training, we train 5 epochs using the Adam~\cite{kingma2014adam} optimizer with a learning rate of 0.0001, and the batch size is set to 16. For the detail settings about FL, the communication round $T$ is set to 50, and the fraction of active clients in each round is designed to achieve a level of participation equivalent to 50 clients, thus ensuring the data distribution is representative of the overall population. Besides, we observed that the statistics of FLAIR~\cite{song2022flair} suggest that the number of images across different users may have significant variations, leading to severe quantity skew issues. Thus, we follow some previous FL works~\cite{li2019convergence,cho2020client} focusing on non-uniform client sampling to handle this issue. The concept is sampling clients at random such that the probability is the corresponding fraction of data at that client. With the help of this sampling scheme, we could guarantee our client quality, avoiding biased to very tiny local client datasets. Due to page limitations, we report the results with different sampling strategies in the \textit{supplementary materials}. For all experiments, we implement our model by PyTorch and conduct training on a single NVIDIA RTX 3090Ti GPU with 24GB memory.

\begin{table}[t]
  \centering
  \resizebox{1.0\columnwidth}{!}{
  \begin{tabular}{@{}lcccccccc@{}}
    \toprule
    Metrics & C-AP & C-P & C-R & C-F1 & O-AP & O-P & O-R & O-F1\\
    \midrule
    \multicolumn{8}{@{}l@{}}{\textit{\textbf{MS-COCO}}} \\
    FedAvg & 69.20  & 71.00  & 60.30  & 65.20  & 77.80  & 75.80  & 65.30  & 70.20  \\
    FedC-Tran & 76.70  & 76.00  & 67.10  & 71.20  & 83.90  & 79.40  & 71.60  & 75.30  \\
    Ours & \textbf{78.30} & \textbf{77.20} & \textbf{70.00} & \textbf{73.40} & \textbf{84.70} & \textbf{80.20} & \textbf{73.70} & \textbf{76.80} \\
    \midrule
    \multicolumn{8}{@{}l@{}}{\textit{\textbf{PASCAL VOC}}} \\
    FedAvg & 87.50  & 87.90  & 73.30  & 79.90  & 91.80  & 91.70  & 78.30  & 84.50  \\
    FedC-Tran & 89.60  & 88.20  & 79.60  & 83.60  & 93.70  & 91.70  & 83.40  & 87.30  \\
    Ours & \textbf{90.80}  & \textbf{88.80}  & \textbf{82.50}  & \textbf{85.50}  & \textbf{94.10}  & \textbf{91.80}  & \textbf{85.30}  & \textbf{88.40}  \\
    \bottomrule
  \end{tabular}}
  \caption{Comparisons on MS-COCO and PASCAL VOC for our FedLGT with FL baselines.}
  \label{tab:table_coco_voc}
\end{table}

\subsection{Performance Results and Analysis}
As for the evaluation metrics, we follow the convention of works on multi-label image
classification~\cite{lanchantin2020general,ridnik2021asymmetric} and use the metrics of per-class (C) and overall (O) average precision (i.e., C-AP and O-AP), precision (C-P, O-P), recall (C-R, O-R), F1 scores (C-F1, O-F1), with details provided in the \textit{supplementary materials}.
For the coarse-grained task in FLAIR, we first utilize C-Tran~\cite{lanchantin2020general} to conduct centralized learning experiments and compare the centralized results to pure ResNet-18 baseline. As shown in ~\Cref{tab:table_1}, the centralized C-Tran improves the performance. However, as we transfer the C-Tran directly to FL (i.e., FedC-Tran across all our tables), it surpasses the baseline FedAvg though. The gap between FedC-Tran and centralized is still present and has room for improvement. Hence, our proposed method could perform favorably against FedC-Tran (e.g., $4.9\%$ on C-AP, $12\%$ on C-F1, etc.). Similarly, as for the fine-grained task shown in~\Cref{tab:table_2}, the centralized C-Tran still overcomes ResNet-18 baseline. But, FedC-Tran has a huge performance drop when transferred to FL compared to coarse-grained ones, which may be caused by the corrupted label relationships due to the aggregation step in FL. Thus, our method could avoid corrupt label relationships, leading to much more performance boosting (e.g., over $3\times$ on C-AP, $1.5\times$ on C-F1, etc.). Besides, as~\Cref{tab:table_coco_voc} presented, we also verify our effectiveness on extra datasets designed for centralized multi-label problems (MS-COCO, PASCAL VOC), and still achieve favorable performance over FedC-Tran baseline.

\begin{table}[t]
  \centering
  \resizebox{0.8\columnwidth}{!}{
  \begin{tabular}{@{}lcccc@{}}
    \toprule
    Metrics & C-AP & C-F1 & O-AP & O-F1\\
    \midrule
    FedC-Tran & 56.00 & 43.10 & 88.10 & 77.90  \\
    FedC-Tran + \textit{CA-MLE} & 56.10 & 45.00 & 88.30 & 78.40  \\
    FedC-Tran + \textit{ULE} & 59.70 & 54.90 & 88.30 & 78.90 \\
    \midrule
    Ours & \textbf{60.90} & \textbf{55.10} & \textbf{88.70} & \textbf{79.70} \\
    \bottomrule
  \end{tabular}}
  \caption{Abaltion studies of our FedLGT using coarse-grained task on FLAIR. Note that \textit{CA-MLE} means client-aware masked label embedding, while \textit{ULE} is universal label embedding. \textbf{Bold} denotes the best result.}
  \label{tab:table_abs}
\end{table}

\subsection{Ablation Studies}
As reported in~\Cref{tab:table_abs}, we first perform ablation studies about the two components in our method. From ``FedC-Tran +\textit{ CA-MLE}" of~\Cref{tab:table_abs} (i.e., only using the client-aware masked label embedding), we observe that the improvement is not obvious (e.g., $0.1\%$ on C-AP) compared to FedC-Tran baseline, which may be caused by the corrupted label correlations. Next, from ``FedC-Tran + \textit{ULE}" of~\Cref{tab:table_abs} (i.e., only using universal label embedding), the improvement (e.g., $3.7\%$ on C-AP) becomes more significant since the label embeddings are not corrupted by aggregation step. Thus, it can be seen the importance of ULE for the robustness and semantically label correlations. In the last row, our proposed method including the two aforementioned components is able to achieve satisfactory performance compared with the baselines (e.g., $4.9\%$ on C-AP), and highlights the importance of ULE with the further improvements introduced by CA-MLE. Due to page limitation, additional ablation studies on the fine-grained FLAIR are provided in the \textit{supplementary materials}.

\section{Conclusion} \label{sec:conclusion}
In this paper, we tackle the challenging problems of multi-label FL. With the proposed framework of FedLGT, we are able to exploit local label correlation via learning client-aware masked label embedding. With universal label embedding derived from pre-trained vision and language model, the alignment of locally learned models can be performed in the same embedding space, allowing aggregation of such models for improved performance. We verified our proposed methods by conducting extensive experiments on the challenging dataset FLAIR as well as benchmarks of MS-COCO and PASCAL VOC. Our experiments confirmed the robustness and effectiveness of our proposed learning scheme for multi-label FL.

\section*{Acknowledgements}
This work is supported in part by the National Science and Technology Council under grant 112-2634-F-002-007.  We also thank to National Center for High-performance Computing (NCHC) for providing computational and storage resources.

\bibliography{aaai24}

\clearpage
\appendix
\renewcommand\thesection{\Alph{section}}
\renewcommand{\thefigure}{A\arabic{figure}}
\renewcommand{\thetable}{A\arabic{table}}
\setcounter{figure}{0}   
\setcounter{table}{0}   
\setcounter{section}{0}

\section{Evaluation Metrics}
Following the convention of works on multi-label image classification~\cite{chen2019multi,chen2019learning,chen2020knowledge,lanchantin2020general,ridnik2021asymmetric}, the label is predicted present if the probability is larger than 0.5. Besides, by the convention from previous works, we report the metrics in per-class (C) and overall (O) types, including the average precision (C-AP, O-AP), precision (C-P, O-P), recall (C-R, O-R), F1 scores(C-F1, O-F1). The formulation of metrics is shown below:
\begin{equation}\label{eq:metric_perclass}
\begin{aligned}
    &\text{C-P}=\frac{1}{C} \sum_{i} \frac{M_{c}^{i}}{M_{p}^{i}}&&\text{O-P}=\frac{\sum_{i} M_{c}^{i}}{\sum_{i} M_{p}^{i}} \\
    &\text{C-R}=\frac{1}{C} \sum_{i} \frac{M_{c}^{i}}{M_{g}^{i}} &&\text{O-R}=\frac{\sum_{i} M_{c}^{i}}{\sum_{i} M_{g}^{i}}\\
    &\text{C-F1} = \frac{2 \times \text{C-P} \times \text{C-R}}{\text{C-P} + \text{C-R}}&&\text{O-F1} = \frac{2 \times \text{O-P} \times \text{O-R}}{\text{O-P} + \text{O-R}},\\
\end{aligned}
\end{equation}
where $C$ is the total number of labels, $M_c^i$ is the number images predicted correctly for $i$-th class, $M_{p}^{i}$ is the number images predicted for $i$-th class, and $M_{g}^{i}$ is the number of ground truth images for $i$-th class. Since the precision and recall results may be influenced by the chosen threshold for decisions about positive/negative classes, C-F1, O-F1, C-AP, and O-AP are crucial as they provide a more comprehensive view of model performance generally.

\section{More Ablation Studies and Experiments}
\subsection{Parameters Analysis for CA-MLE}
In our CA-MLE module, we have to adjust the uncertainty margin (i.e., $\varepsilon$) to determine where should be masked or not. Thus, we ablate various different settings of uncertainty margin. As~\Cref{tab:table_abs_1} shown, we could observe that if the $\varepsilon$ is smaller the performance is enhanced more, which is reasonable since focusing more on the classes that indeed are uncertain for the model during training is crucial to our design.

\begin{table}[h]
  \centering
  \resizebox{0.8\columnwidth}{!}{
  \begin{tabular}{@{}lcccc@{}}
    \toprule
    Metrics & C-AP & C-F1 & O-AP & O-F1\\
    \midrule
    $\varepsilon=0.05$ & 59.10 & 53.00 & 87.70 & 77.70  \\
    $\varepsilon=0.03$ & 59.30 & 52.80 & 87.80 & 77.80  \\
    $\varepsilon=0.02$ & \textbf{60.90} & \textbf{55.10} & \textbf{88.70} & \textbf{79.70} \\
    \bottomrule
  \end{tabular}}
  \caption{Ablations on uncertainty margin for CA-MLE. \textbf{Bold} denotes the best result.}
  \vspace{-2mm}
  \label{tab:table_abs_1}
\end{table}

\subsection{Choices of Label Embedding}

\begin{table}[t]
  \centering
  \resizebox{0.8\columnwidth}{!}{
  \begin{tabular}{@{}lcccc@{}}
    \toprule
    Metrics & C-AP & C-F1 & O-AP & O-F1\\
    \midrule
    GloVe & 57.90  & 48.90  & 86.70  & 75.90  \\
    BERT & 58.90  & 50.70  & 87.80  & 77.40 \\
    CLIP & \textbf{60.90} & \textbf{55.10} & \textbf{88.70} & \textbf{79.70} \\
    \bottomrule
  \end{tabular}}
  \caption{Different text encoders for building ULE. \textbf{Bold} denotes the best result.}
  \label{tab:table_abs_2}
\end{table}

In our main paper, we already show that with the help of a frozen CLIP~\cite{radford2021learning} text encoder, we could obtain improved model performance compared to original learnable label embeddings under FL. As shown in~\Cref{tab:table_abs_2}, we ablate different text encoders to build ULE to verify our effectiveness. Among our comparisons, the CLIP text encoder achieves the most satisfactory performance, while the BERT~\cite{devlin2018bert} and GloVe~\cite{pennington2014glove} perform slightly worse than CLIP embeddings. The results reveal that the cross-modal pre-trained model (i.e., CLIP) may learn more about the relationships between images and text at the pre-trained stage, thus it has the ability to benefit our multi-label FL tasks more compared to the uni-modal ones (i.e., GloVe, BERT).

\subsection{Discussion of Input Label Embedding}

\begin{table}[h]
  \centering
  \resizebox{0.8\columnwidth}{!}{
  \begin{tabular}{@{}lcccc@{}}
    \toprule
    Metrics & C-AP & C-F1 & O-AP & O-F1\\
    \midrule
    Avg. prompt & 60.30  & 52.60 & 88.60 & 79.10  \\
    Concat. prompt & \textbf{60.90} & \textbf{55.10} & \textbf{88.70} & \textbf{79.70} \\
    \bottomrule
  \end{tabular}}
  \caption{Different strategies to build ULE for coarse-grained FLAIR. \textbf{Bold} denotes the best result.}
  \label{tab:table_abs_3}
\end{table}

Since the classes are abstract for FLAIR coarse-grained level, we build the prompts in two ways via the fine-grained level information. The first is concatenating all the fine-grained labels that belong to each coarse-grained label to build the prompts (i.e., \textbf{Concat. prompt} in~\Cref{tab:table_abs_3}). The second would build all text embeddings of the fine-grained labels, taking the average on corresponding fine-grained ones to build for each coarse-grained label (i.e., \textbf{Avg. prompt} in the table). As shown in~\Cref{tab:table_abs_3}, we choose to build the embeddings by \textbf{Concat. prompt} to achieve better performance.

\begin{table*}[t]
  \centering
  \resizebox{2.0\columnwidth}{!}{
  \begin{tabular}{@{}lccccccccccccccccc@{}}
    \toprule
     & structure & equipment & material & outdoor & plant & food & animal & liquid & art & interior room & light & recreation & celebration & fire & music & games & religion \\
    \midrule
    FedC-Tran & 91.10  & 93.52  & 68.53  & 95.46  & 92.91  & 95.46  & 91.90  & 79.81  & 46.20  & 68.53  & 34.22  & 19.55  & 23.89  & 50.22  & 9.12  & 7.88  & 3.07  \\
    Ours & \textbf{91.20}  & 93.40  & \textbf{69.10}  & \textbf{95.50}  & \textbf{93.10}  & \textbf{95.60}  & \textbf{91.90}  & \textbf{80.20}  & \textbf{48.80}  & \textbf{68.80}  & \textbf{35.90}  & \textbf{19.60}  & \textbf{35.20}  & \textbf{61.60}  & \textbf{20.40}  & \textbf{20.00}  & \textbf{8.40}  \\
    \bottomrule
  \end{tabular}}
  \caption{Averaged precision (AP) for each coarse-grained class on FLAIR. Note the columns are sorted from left to right according to the number of samples for each class. \textbf{Bold} denotes the best result.}
  \label{tab:table_per_class}
\end{table*}

\subsection{Per-class Results on FLAIR}
We show the per-class averaged precision to analyze the details in coarse-grained level FLAIR.
As shown in the~\Cref{tab:table_per_class}, we obtain the per-class results on the FLAIR~\cite{song2022flair} test set under coarse-grained level, and the table columns are sorted according to the number of samples. Due to severe label skewness in FLAIR, the classes that have fewer samples may tend to be misclassified. Compared our work to the FedC-Tran, it can be seen that our method performs even better on the classes with fewer samples. For example, for the class ``games", we achieve $12.12\%$ improvement, and $5.33\%$ for the class with the least number of samples (i.e., ``religion"). 

Besides, for a more practical scenario, consider the labels of ``light", ``fire", and ``games" in the scenes of shooting games and candle decoration. The above label correlation would be expected to be very different in different scenes. As shown in~\Cref{tab:table_per_class}, compared to FedC-Tran, our method achieved improved per-class results for these 3 classes (i.e., $\textbf{1.7\%}$, $\textbf{11.38\%}$, and $\textbf{12.12\%}$, respectively), showing the effectiveness of our proposed learning scheme.

\subsection{Different Client Sampling Strategies}

\begin{table}[h]
  \centering
  \resizebox{0.8\columnwidth}{!}{
  \begin{tabular}{@{}lrrrr@{}}
    \toprule
    Metrics & C-AP & C-F1 & O-AP & O-F1\\
    \midrule
    \multicolumn{5}{@{}l@{}}{\textit{\textbf{Uniform client sampling}}} \\
    FedC-Tran & 1.80  & 0.70  & 25.40  & 23.70  \\
    Ours & 4.00 & 1.70 & 30.40 & 30.80  \\
    \midrule
    \multicolumn{5}{@{}l@{}}{\textit{\textbf{Non-uniform client sampling}}} \\
    FedC-Tran & 3.30 & 1.50  & 36.70 & 31.70  \\
    Ours & \textbf{10.60} & \textbf{2.30} & \textbf{42.20} & \textbf{33.40} \\
    \bottomrule
  \end{tabular}}
  \caption{Comparison of different client sampling strategies on the fine-grained task of FLAIR. \textbf{Bold} denotes the best result.}
  \label{tab:table_sample_fg}
\end{table}

We follow some previous FL works~\cite{li2019convergence,cho2020client} focusing on non-uniform client sampling to alleviate the severe quantity skew issues caused by inherent statistics of FLAIR~\cite{song2022flair}. With the help of this sampling scheme, we could avoid the model from being biased toward very tiny local client datasets, which may hurt the aggregation step in FL. We ablate different sampling schemes on fine-grained FLAIR to verify the effectiveness of non-uniform sampling to ensure the dataset quality of local clients. As shown in~\Cref{tab:table_sample_fg}, it can be seen that though the performance has been improved when uniform sampling, we can obtain more satisfactory results under non-uniform sampling, which indicates that the importance of the dataset quality of local clients.

\subsection{Ablation Study for Fine-grained FLAIR}

\begin{table}[t]
  \centering
  \resizebox{0.8\columnwidth}{!}{
  \begin{tabular}{@{}lcccc@{}}
    \toprule
    Metrics & C-AP & C-F1 & O-AP & O-F1\\
    \midrule
    FedC-Tran & 3.30 & 1.50 & 36.70 & 31.70  \\
    FedC-Tran + \textit{CA-MLE} & 5.50 & 1.60 & 37.50 & 31.90  \\
    FedC-Tran + \textit{ULE} & 7.20 & 2.00 & 38.50 & 32.60 \\
    \midrule
    Ours & \textbf{10.60} & \textbf{2.30} & \textbf{42.20} & \textbf{33.40} \\
    \bottomrule
  \end{tabular}}
  \caption{Ablation studies of our FedLGT using fine-grained task on FLAIR. Note that \textit{CA-MLE} means client-aware masked label embedding, while \textit{ULE} is universal label embedding. \textbf{Bold} denotes the best result.}
  \label{tab:table_abs_fg}
\end{table}

As shown in~\Cref{tab:table_abs_fg}, we perform ablation studies in fine-grained level FLAIR. From ``FedC-Tran +\textit{ CA-MLE}" of~\Cref{tab:table_abs_fg}, we observe that the improvement is not obvious (e.g., $2.2\%$ on C-AP) compared to FedC-Tran baseline, which may also be caused by the corrupted label correlations, and this corruption could become more serious under fine-grained level. Then, it can be observed that from ``FedC-Tran + \textit{ULE}" of~\Cref{tab:table_abs_fg}, the improvement (e.g., $3.9\%$ on C-AP) becomes more significant, which may be caused by our design of ULE can avoid the label embeddings to be corrupted by FL aggregation step. Thus, it can be seen the importance of ULE for the robustness and semantically label correlations. Finally, our proposed method including the two aforementioned components is able to achieve satisfactory performance compared with the baselines (e.g., $7.3\%$ on C-AP), and points up the effectiveness of ULE with the further improvements introduced by CA-MLE, even under more difficult scenarios (i.e., fine-grained level of FLAIR).

\subsection{Larger Rounds Results on FLAIR}
Due to resource constraints, we perform our main experiments with a communication round of 50 as shown in the main paper. To further verify our robustness with more rounds, we additionally conduct experiments in coarse-grained levels for\textbf{ 500 rounds} as shown in~\cref{tab:table_baseline}, which again confirms our superiority over the baselines.

\begin{table}[h]
  \centering
  \resizebox{0.8\columnwidth}{!}{
  \begin{tabular}{@{}lrrrr@{}}
    \toprule
    Metrics & C-AP & C-F1 & O-AP & O-F1\\
    \midrule
    \multicolumn{5}{@{}l@{}}{\textit{\textbf{50 Rounds}}} \\
    FedAvg & 40.63 & 32.74 & 77.39 & 65.53  \\		
    FedC-Tran & 56.00 & 43.10 & 88.10 & 77.40  \\
    Ours & \textbf{60.90} & \textbf{55.10} & \textbf{88.70} & \textbf{79.70} \\
    \midrule
    \multicolumn{5}{@{}l@{}}{\textit{\textbf{500 Rounds}}} \\
    FedAvg & 50.20 & 43.00 & 85.50 & 76.00  \\		
    FedC-Tran & 59.80 & 52.50 & 88.60 & 79.20  \\
    Ours & \textbf{65.70} & \textbf{62.00} & \textbf{89.90} & \textbf{80.50} \\
    \bottomrule
  \end{tabular}}
  \caption{Comparison of different communication rounds on the coarse-grained task of FLAIR.}
  \label{tab:table_baseline}
\end{table}

\end{document}